\documentclass[letterpaper, 10 pt, conference]{ieeeconf}  

\IEEEoverridecommandlockouts                              

\usepackage{amsmath} 
\usepackage{amssymb}  
\usepackage{tikz}
\usetikzlibrary{positioning, arrows.meta, quotes}
\usetikzlibrary{shapes,snakes}
\usetikzlibrary{bayesnet}
\tikzset{>=latex}
\tikzstyle{plate caption} = [caption, node distance=0, inner sep=0pt,
below left=5pt and 0pt of #1.south]
\usepackage{algorithmic}
\usepackage{graphicx}
\usepackage{subcaption}
\usepackage{changepage}
\usepackage{textcomp}
\usepackage{xcolor}
\usepackage{url}
\usepackage{multirow}
\usepackage[bookmarks=true]{hyperref} 
\usepackage{array}
\usepackage{booktabs}
\usepackage{tabularray}
\usepackage{makecell}
\usepackage{times}
\usepackage{comment}
\usepackage{cite}
\usepackage{xurl}

\newcounter{repofootnote}

\title{Robot Learning from Human Demonstrations: Handwritten\\Alphabet Trajectories and Human-Likeness Evaluation}

\author{Alperen Kenan$^{1}$, Paul Bremner$^{1}$, and Manuel Giuliani$^{2}$
\thanks{$^{\star}$This work was supported by the European Commission’s Marie Skłodowska-Curie Actions (MSCA) Project RAICAM (GA 101072634), and UK Research and Innovation (UKRI) grant number EP/X025977/1.}
\thanks{For the purpose of open access, the author has applied a Creative Commons Attribution (CC BY) licence to any Author Accepted Manuscript version arising.}
\thanks{$^{1}$Alperen Kenan is a PhD student, and Paul Bremner is an Associate Professor in Human-Robot Interaction at the Bristol Robotics Laboratory, University of the West of England, Bristol, United Kingdom. Emails: {\tt\small alperen.kenan@uwe.ac.uk, paul2.bremner@uwe.ac.uk}}%
\thanks{$^{2}$Manuel Giuliani is a Professor for Geriatric Robotics at the Faculty of Electrical Engineering, Kempten University of Applied Sciences, Kempten, Germany. Email: {\tt\small manuel.giuliani@hs-kempten.de}}%
}

\begin{document}

\maketitle
\thispagestyle{empty}
\pagestyle{empty}

\begin{abstract}
Learning from demonstration (LfD) provides a developmental framework through which robots can develop motor skills by observing and imitating human dynamics, reducing reliance on explicit programming to teach a skill to a robot. The resulting human-like robot motion is recognised as a key factor in building trust and enabling natural collaboration in human-robot interaction. This paper presents a framework for learning human-like robot motion from demonstration, including data collection, probabilistic trajectory learning, and perceptual user evaluation. A dataset of 3,142 handwriting demonstrations was collected from 22 participants across all 52 Latin alphabet character-case combinations via a touchscreen teleoperation interface, capturing planar position, contact force, and timing. Building on the widely used Gaussian Mixture Model and Gaussian Mixture Regression approach for learning from demonstration, the framework is extended in this work by incorporating force and normalised time dimensions to enable richer representation of human dynamics, and adapting it to handle non-continuous, multi-segment trajectories, enabling generalisation across demonstrations. A user study with 21 participants evaluated the perceived human-likeness of the generated trajectories using a continuous scale anchored between robotic and human-like motion, normalised to 0-100 where 50 represents the neutral midpoint. The generated trajectories achieved an overall human-likeness score of 71.50 (SD=22.56), indicating that the majority of trajectories were perceived as more human-like. Participants identified geometric positioning and trajectory sequence as the most influential perceptual factors, and reported positive attitudes toward human-like robot behaviour. The datasets are released as open-source, providing a reproducible benchmark for developing and evaluating human-like robot motion methods.

\end{abstract}

\section{Introduction}

The role of robots in industrial and collaborative environments is shifting from isolated task executors to active participants in shared human workspaces \cite{morandini2025collaborative}. As robots take on increasingly complex roles alongside humans, the ability to move in ways that feel natural and predictable to human collaborators has become a critical design consideration \cite{dragan2013legibility}. Motion and trajectory are key channels of non-verbal communication in human-robot interaction (HRI), conveying intent, readiness, and social cues that influence how humans perceive and respond to robots \cite{Saunderson2019, Kim2025}. Robots that move in human-like ways have been shown to be more readily accepted, more trusted, and more effective collaborators than those executing rigid, pre-programmed motions \cite{Kim2025, Dragan2015}.

A promising approach to achieving human-like robot motion is Learning from Demonstration (LfD), also referred to as Programming by Demonstration (PbD), in which a robot learns to reproduce a task by observing human examples rather than through explicit programming \cite{Argall2009, Ravichandar2020}. This paradigm is particularly attractive because it enables non-expert users to teach robots new behaviours intuitively, without requiring knowledge of robotics or software  engineering \cite{Ehrenmann2002, Bautista2014}. However, generating trajectories that  are not only kinematically correct but also perceived as human-like by observers remains  an open challenge, as most existing LfD approaches focus on spatial accuracy and task completion, with limited attention to the human dynamics of motion \cite{Correia2024}.

Despite growing interest in this area, several gaps remain in the literature. Few publicly available datasets capture the full dynamics of human demonstrations, including contact force and timing alongside position, limiting reproducibility, cross-method benchmarking, and pushing most existing approaches to model spatial trajectories alone while overlooking force and temporal dynamics. Furthermore, human-centred evaluations of perceived human-likeness are rarely conducted, leaving it unclear whether learned motions are actually perceived as natural by human observers and whether they are suitable for real-world HRI applications.

\begin{figure}[t]
\centering
\includegraphics[width=1\columnwidth]{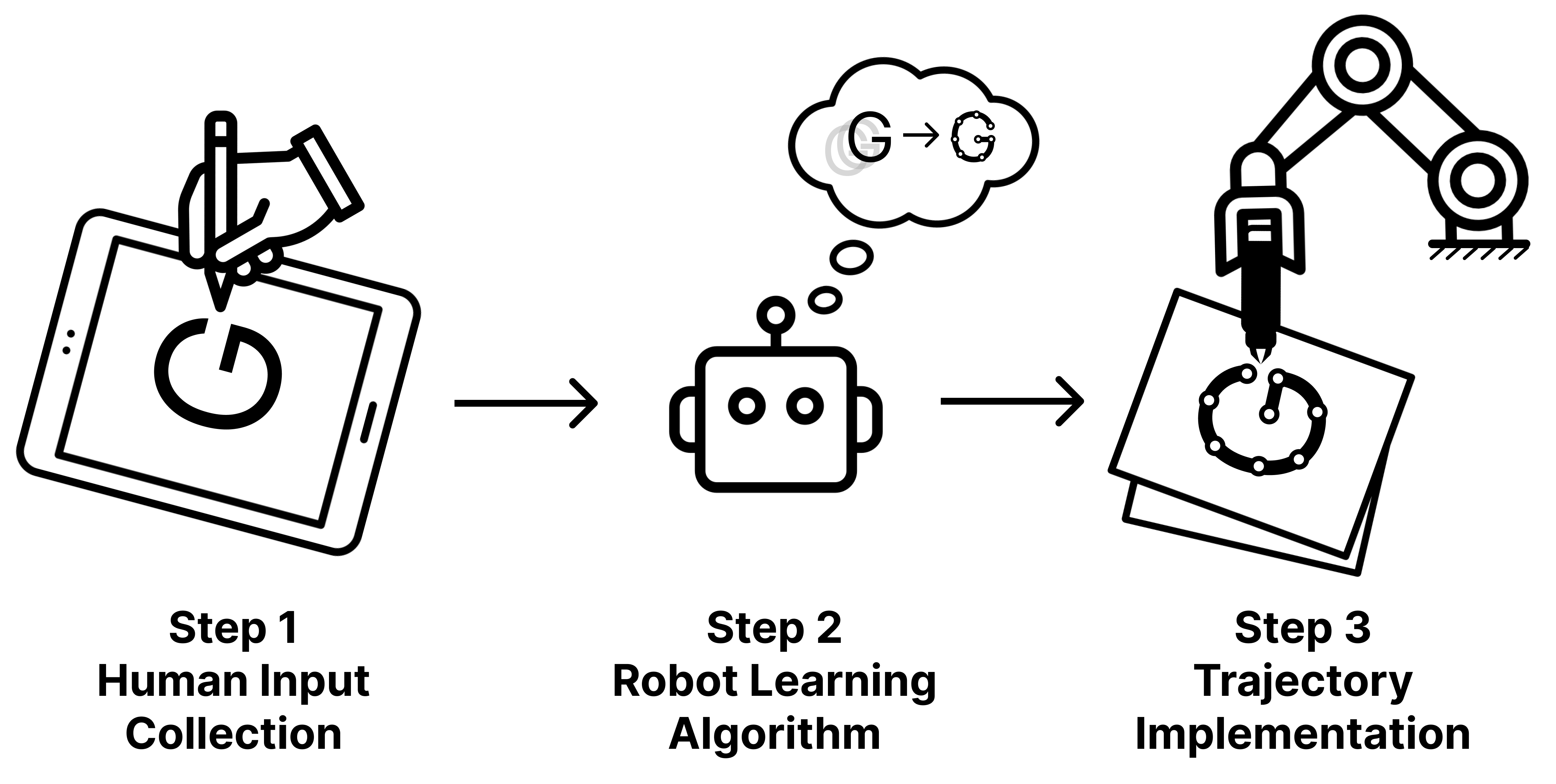}
\caption{Teaching-by-demonstration pipeline illustration}
\label{fig:illustration}
\vspace{-15pt}
\end{figure}

To address these gaps, this paper presents a complete framework for collecting human motion demonstrations, training a robot learning algorithm, and evaluating the perceived human-likeness of the generated trajectories. Human handwriting was chosen as the target task due to its complexity in spatial, temporal, and force dynamics, its multi-segment structure, and its well-defined character representations. A dataset of 3,142 demonstrations was collected from 22 participants across all 52 Latin alphabet character-case combinations via a touchscreen teleoperation interface, capturing position, timing, and contact force. Building on the Gaussian Mixture Model and Gaussian Mixture Regression (GMM+GMR) framework presented in \cite{Calinon2010}, this work extends it with force and normalised time dimensions, and adapts it to handle non-continuous multi-segment trajectories, a limitation not addressed by standard GMM+GMR formulations. This approach allows generalisation across demonstrations rather than replaying any single trial, reflecting a learning process in which experience gradually leads to more refined behaviour. A user study with 21 participants provided a human-centred assessment of the perceived human-likeness of the generated motions using a simulated robot.

This paper makes three primary contributions: (i) an open-source human handwriting dataset capturing position, timing, and force data across all 52 Latin alphabet character-case combinations, offering a resource for researchers to benchmark their own learning methods; (ii) two extensions to the GMM+GMR framework: incorporation of contact force and normalised time dimensions, and a multi-segment approach handling trajectory discontinuities from pen lifts; and (iii) a user-centred evaluation of the generated trajectories using a simulated robot, providing an assessment of the perceived human-likeness of the learned motions. Figure \ref{fig:illustration} illustrates the three main steps of the framework.

The remainder of the paper is organised as follows. Section~\ref{sec:relatedwork} reviews prior work on related topics. Section~\ref{sec:Methodology} describes the experimental methodology, including the first user study for human demonstration data collection and the second user study for evaluating the perceived human-likeness of the generated trajectories. Section~\ref{sec:teleoperation} details the GMM+GMR learning algorithm used to generate robot motion from the collected dataset and the teleoperation setup. Section~\ref{sec:Results} presents the statistical properties of both datasets and the user evaluation results. Section~\ref{sec:Discussion} interprets the results, discusses the contributions and implications of the research, and outlines limitations and future work. Finally, Section~\ref{sec:Conclusions} summarises the main findings and their broader significance for robot learning from demonstration applications.

\section{Related Work}
\label{sec:relatedwork}

\subsection{Human Handwriting Datasets}

Several datasets have been proposed to study human handwriting and drawing behaviour. Early work such as the IAM handwriting database and more recent datasets like EMNIST captured handwritten characters in standardised image-based formats for recognition tasks \cite{Marti2002,Cohen2017}, whereas more recent efforts have incorporated temporal and kinematic information \cite{Ott2020,AlShamaileh2019}.

However, most existing datasets remain image-based, and those that include dynamic information primarily capture position and stroke timing, with limited consideration of contact force. While datasets such as OnHW \cite{Ott2020} include additional sensing modalities, they are typically collected in unconstrained settings where participants write freely on paper without guidance on character structure, scale, or stroke order, with sensors capturing data throughout the process. This results in high variability, making such datasets less suitable for direct use in robot learning pipelines or for producing consistent trajectories aligned with robotic control requirements. Consequently, their applicability for learning human-like motion in physical interaction tasks is limited.

In contrast, in this work teleoperation of a simulated robot is used during data collection, incorporating realistic constraints such as a maximum end-effector speed, requiring participants to adjust their trajectories accordingly. In addition, predefined character templates, font size, and stroke order are guided through the interface. This yields structured and consistent demonstrations that are better suited for training and evaluating learning methods for generating robot trajectories.

\subsection{Robot Learning from Demonstration with GMM+GMR}

LfD enables robots to acquire new skills by observing human demonstrations rather than relying on explicit programming \cite{Argall2009, Ravichandar2020, Braglia2025}. This approach lowers the barrier to robot programming, allowing non-expert users to intuitively teach robots new tasks \cite{Ehrenmann2002, Bautista2014}.

Gaussian Mixture Models (GMM), combined with Gaussian Mixture Regression (GMR), have emerged as a widely adopted probabilistic framework for encoding and reproducing motion from demonstrations \cite{Hewitt2017, Calinon2012}. GMM encodes the joint distribution of demonstrated trajectories, while GMR retrieves a smooth, generalised trajectory conditioned on a query variable such as time. Calinon et al. \cite{Calinon2010} demonstrated the effectiveness of this approach for learning robot manipulation tasks from multiple demonstrations, showing strong generalisation across variability in the training data. Extensions to task-parameterised formulations \cite{calinon2010learning} and higher-dimensional inputs have further broadened the applicability of the framework, although relatively few works explicitly incorporate contact force as a learning dimension.

More recent imitation learning approaches, including diffusion-based policies \cite{chi2025diffusion} and transformer-based action sequence models \cite{zhao2023learning}, can model temporally extended action sequences and leverage larger demonstration datasets. However, these methods typically require more data and computational resources and are generally less interpretable than probabilistic trajectory models. For the constrained, low-data setting considered in this work, GMM+GMR offers a practical balance of expressiveness, interpretability, computational efficiency, and ease of implementation.

In most existing GMM+GMR-based LfD approaches, the input space is primarily limited to spatial position and time, and trajectories are assumed to be continuous. This study extends both aspects by incorporating force and normalized time as additional dimensions, enabling a richer representation of human dynamics and greater flexibility for integrating further modalities, and by adapting the framework to non-continuous, multi-segment trajectories. Moreover, data are often collected either via teleoperation or manually guiding the robot’s end-effector, both of which introduce robot dynamics into the recorded trajectories, making it challenging to isolate purely human-generated motion. 

\subsection{Human-Like Robot Motion and User Evaluation}

The study of human-like robot motion has received increasing attention in both human–robot interaction (HRI) and developmental robotics. Prior work has shown that robots exhibiting natural, human-like motion are perceived as more predictable, more trustworthy, and easier to collaborate with \cite{Kim2025, Dragan2015}. One indirect method of improving the human-likeness of a robot is to incorporate biological motion principles, such as the two-thirds power law relating speed and curvature in human movement, which have been applied to robot trajectory generation to improve perceived naturalness \cite{Braglia2025, Lacquaniti1983}. More recent approaches have explored imitation learning and deep learning methods for generating expressive robot motion \cite{Ravichandar2020}. 

Human-centred evaluation of robot motion remains relatively underexplored compared to objective performance metrics. Perceptual studies typically assess human-likeness using Likert scales, pairwise comparisons, and the Godspeed questionnaire \cite{Bartneck2009}, and indicate that perceived naturalness depends not only on spatial trajectories but also on motion dynamics such as velocity profiles and force application \cite{Kim2025, Dragan2015}. User evaluations have further highlighted substantial inter-individual variability in perception \cite{Coates2008}.

This work captures human motion dynamics from demonstrations across all 52 uppercase and lowercase alphabet characters, and presents a user-centred evaluation of how closely the learned handwriting trajectories align with human perception of motion dynamics.

\section{Methodology}
\label{sec:Methodology}

Two separate user studies were conducted. The first collected human writing trajectory data while participants teleoperated a simulated robot, capturing position, speed, and force. The second evaluated whether the robot trained on this data produces human-like motion compared to the original fonts. The characters chosen were the 26 Latin alphabet characters in both uppercase and lowercase versions, that are highly nonlinear and complex, with discontinuities that make them challenging to program manually.

\subsection{Participants}
Two separate participant groups were recruited for each study phase to ensure evaluators had no prior knowledge that the robot motion was trained from human demonstrations. In total, 43 participants took part across both studies, with 22 participants contributing to the demonstration data collection phase and 21 participants evaluating the human-likeness of the resulting robot motion.

For the first user study, the 22 participants comprised 14 male and 8 female, aged 22--57 (\emph{M} = 31.45, \emph{SD} = 10.60). Of these, 21 were right-handed and 1 left-handed. Regarding stylus experience, 11 reported using one occasionally, 6 regularly, 4 never, and 1 only once or twice. For the second user study, the 21 participants comprised 14 male and 7 female, aged 21--53 (\emph{M} = 27.24, \emph{SD} = 8.47).

\subsection{Study Environment and Context}
Similarly to the participant groups, the two user studies were conducted in two different laboratories. The first study was conducted at Bristol Robotics Laboratory (Bristol, UK), while the second was conducted at the Autonomous Systems and Robotics Lab, ENSTA, Institut Polytechnique de Paris, Paris, France. This geographical separation helped minimise participant bias that could arise from prior knowledge of how the robot motion was generated.

Before each experiment session, participants completed a pre-experiment questionnaire. In the first study, this included questions on hand dominance and experience with a touchscreen stylus. The second study included a post-experiment questionnaire asking participants about their decision-making process, specifically how important they considered the position, speed, sequence, and force exerted by the robot for their decision making, as well as questions regarding how human-like robot motion affects predictability, trust, and acceptance. Each session began with a study introduction, followed by a brief tutorial and a familiarisation period. In both studies, a UR10 manipulator was simulated in the game engine Unity\footnote{Unity is a real-time 3D development platform used for simulation, visualisation, and interactive applications. See \url{https://unity.com/}.}.

\begin{figure}[h]
\centering
\begin{subfigure}[t]{0.328\columnwidth}
    \includegraphics[width=\linewidth]{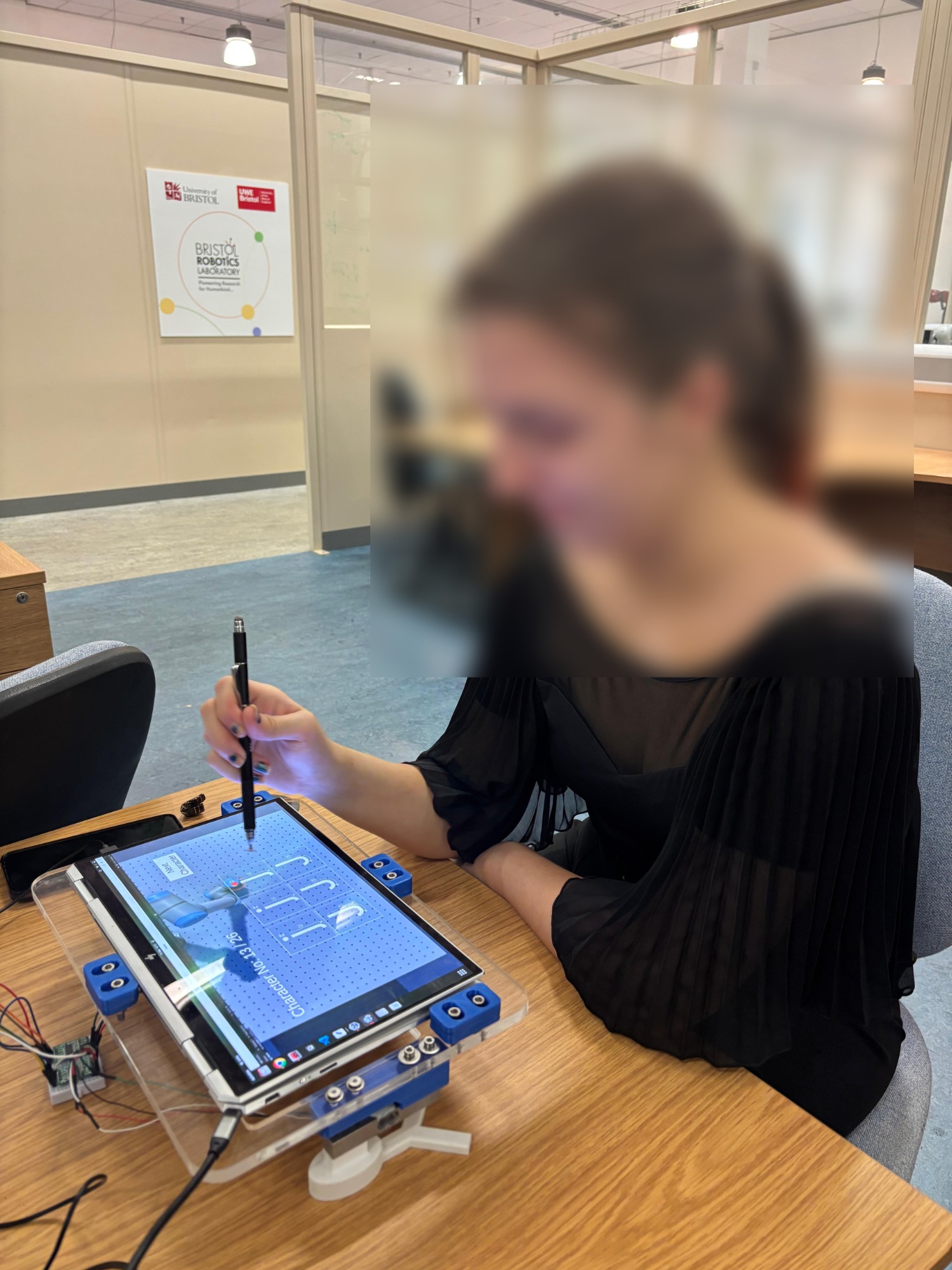}
    \caption{A participant teleoperating the simulated robot.}
    \label{fig:participant}
\end{subfigure}\hspace{0.004\columnwidth}
\begin{subfigure}[t]{0.655\columnwidth}
    \includegraphics[width=\linewidth]{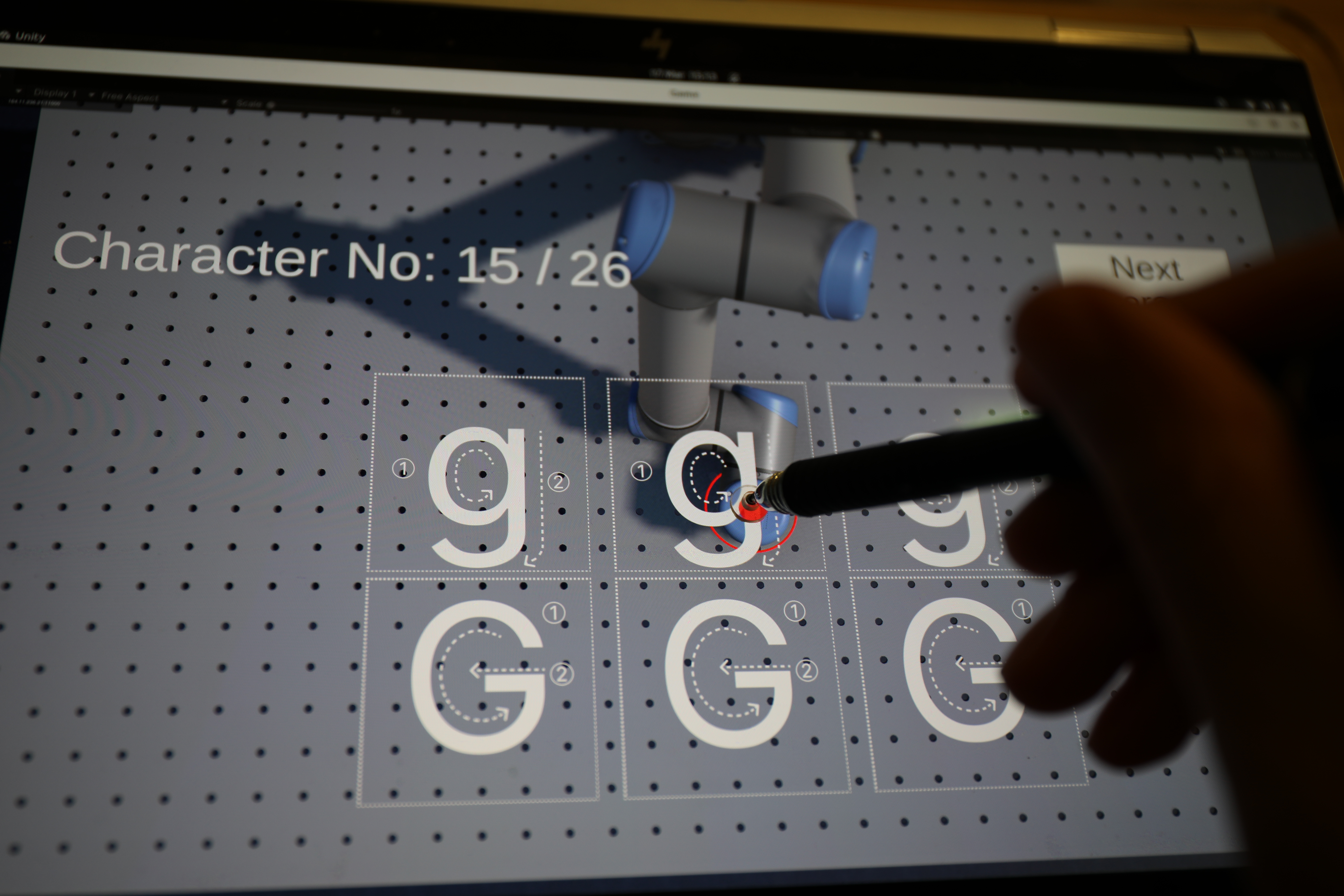}
    \caption{Close-up view of the touchscreen experiment setup.}
    \label{fig:screen}
\end{subfigure}
\caption{Experiment setup for the first user study.}
\label{fig:study1_setup}
\end{figure}

In the first study, participants controlled a simulated robot via a touchscreen teleoperation interface \cite{garcia2026touchscreen}, following different character paths. Following the practice period, they were asked to trace the paths of characters presented in a randomised order to eliminate ordering bias. Each of the 22 participants repeated the motion three times per character for both uppercase and lowercase versions, resulting in 52 characters repeated three times per participant, with each session lasting approximately 10-15 minutes. The touchscreen-based interface was implemented on a 13-inch HP Spectre x360 Convertible laptop with an ELAN2514 touch controller. A capacitive disc-tip stylus pen was used to interact with the touchscreen, allowing participants to hold it like a regular pen without obstructing the interaction area. Two HX711 ADC module load cells with an 80~Hz sampling rate were placed beneath the touchscreen to measure interaction force, and a ROS-based system recorded position, force, and timestamp data. Figure \ref{fig:study1_setup} displays the experimental setup used in the first phase of the study.

\begin{figure}[h]
\centering
\begin{subfigure}[t]{0.348\columnwidth}
    \includegraphics[width=\linewidth]{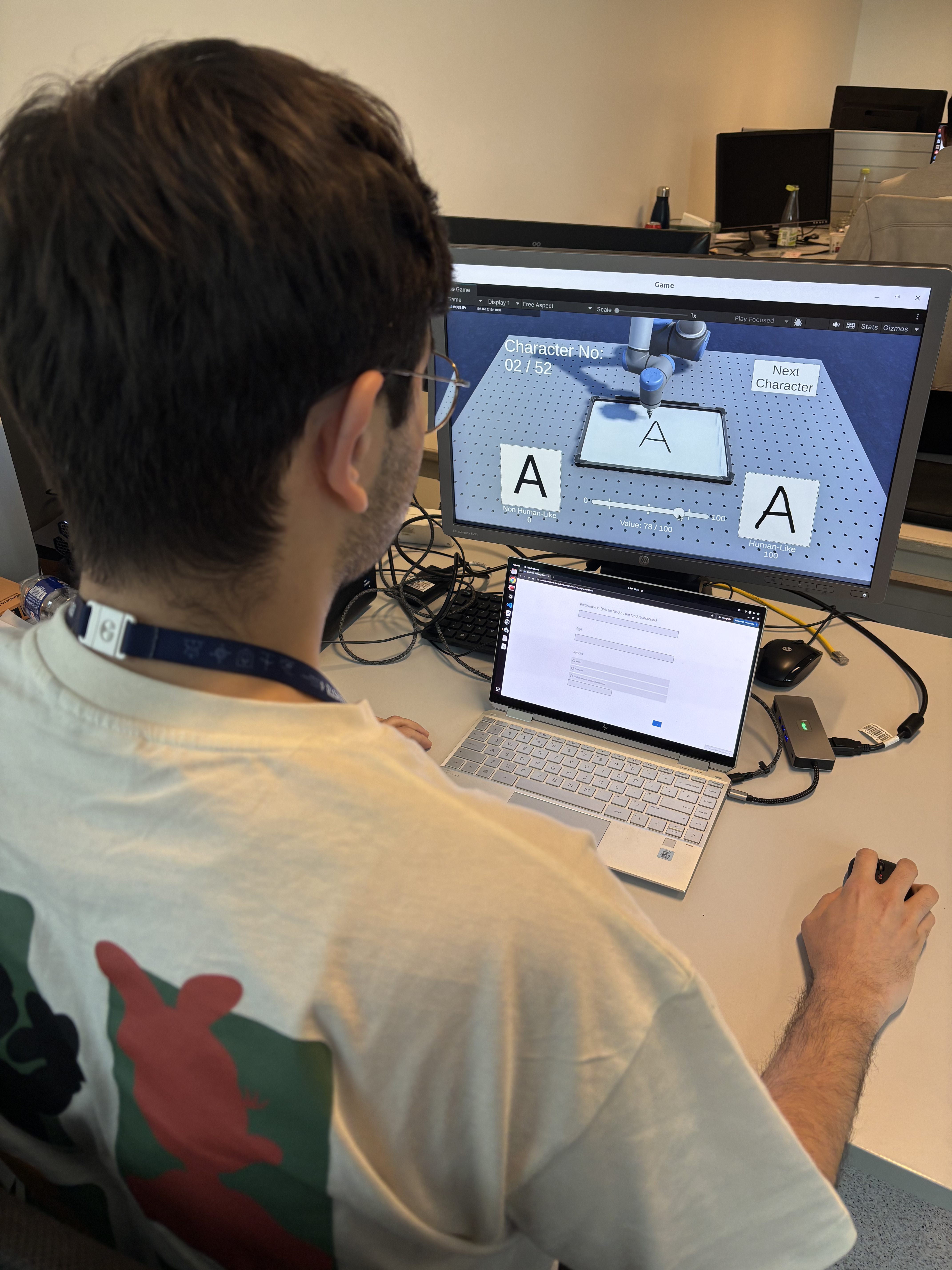}
    \caption{A participant observing the robot motion.}
    \label{fig:study2_a}
\end{subfigure}\hspace{0.002\columnwidth}
\begin{subfigure}[t]{0.628\columnwidth}
    \includegraphics[width=\linewidth]{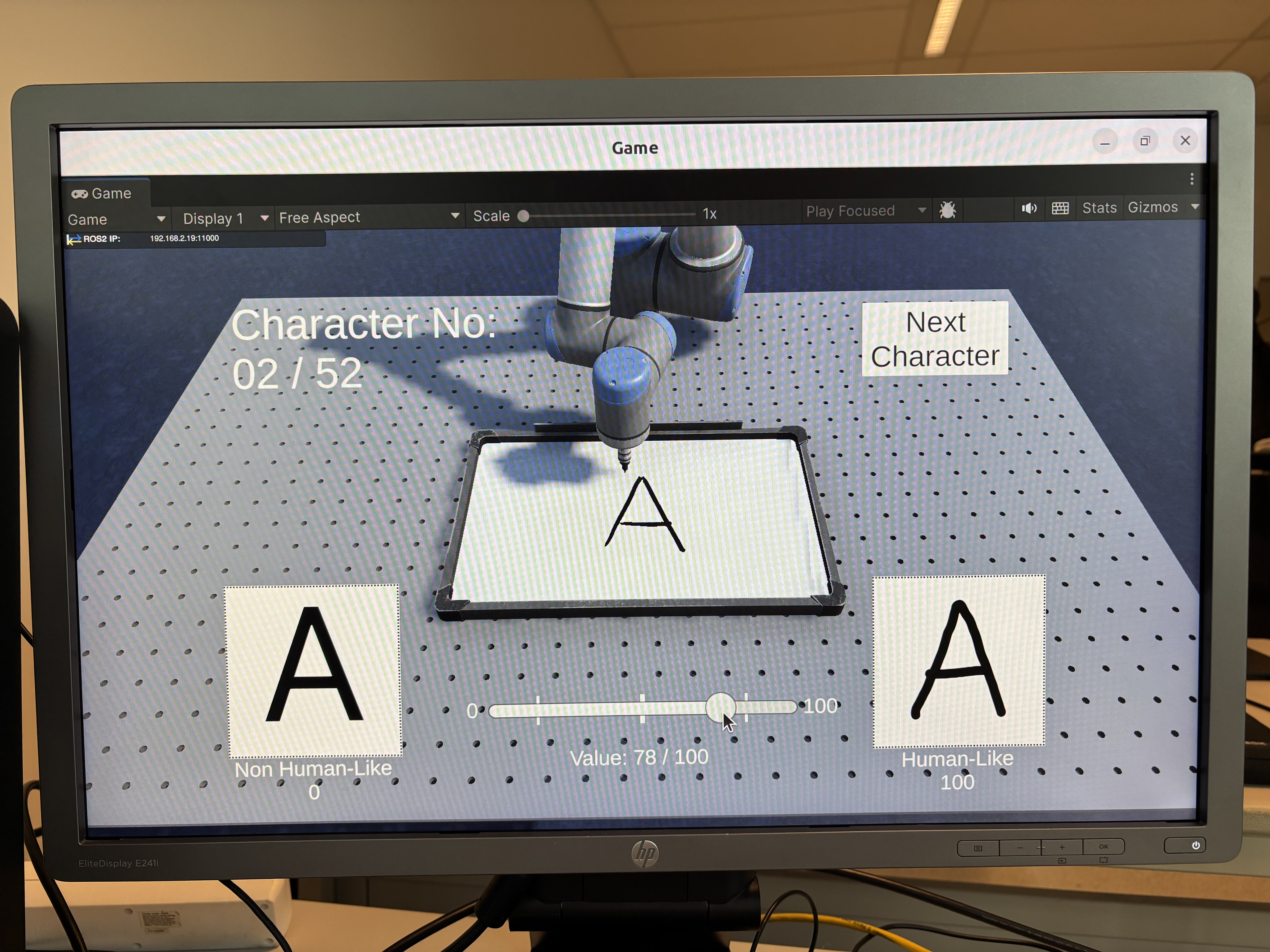}
    \caption{Close-up view of the display.}
    \label{fig:study2_b}
\end{subfigure}
\caption{Experiment setup for the second user study.}
\label{fig:study2_setup}
\end{figure}

In the second study, participants observed a simulated robot drawing characters in randomised order on a 27-inch display. A marker attached to the manipulator reproduced positioning, speed, sequence, and force from the learned demonstrations, with applied pressure reflected through marker thickness and transparency. Participants were instructed to evaluate a combination of the motion trajectory and the final written character appearance, rating the human-likeness of each character on a continuous scale, where the left end represented the original font and the right end human writing. Figure \ref{fig:study2_setup} displays the experimental setup used in the second phase of the study. The open-source code and data for this project are publicly available.\footnote{\label{fn:repo}The repository can be accessed at: \url{https://github.com/kenanalperen/Trajectory-and-Force-Data-for-Handwritten-Alphabet-Generation.git}. It includes the human input character dataset, the algorithm used for the robot learning implementation, and the generated character dataset produced from the training.}
\setcounter{repofootnote}{\value{footnote}}

\subsection{Ethics}
Both phases of the study were approved by the Research Ethics Committee of the University of the West of England (Reference: 13470965). A participant information sheet was provided to all participants prior to obtaining their consent to take part, and signed consent forms were collected before the experiments began. Appropriate measures were taken to ensure participant confidentiality and data security throughout the study. Participants retained the right to withdraw at any point and to request the removal of their data up to seven days following their participation.

\section{Robot Learning Algorithm}
\label{sec:teleoperation}

\subsection{Human Input Dataset}

Recorded position data is used to categorise characters by their region, and the corresponding force values are filtered such that any data point below a threshold of 0.1\,N is rejected to remove recordings where there was no pen contact with the surface. For each recording, the absolute timestamp is adjusted so that each character instance starts from zero. Each recording was then manually inspected to remove missed letters, characters with incorrect positioning, wrong trajectory order, and similar errors, ensuring the remaining data is suitable for training the robot motion. Of the initial 3,424 character instances, 282 (8.24\%) were excluded due to being unfit for training, leaving 3,142 character instances (an average of 60.4 per character) for use in the final dataset. This dataset is openly available in the shared repository\footnotemark[\value{repofootnote}] 
and other researchers are encouraged to use it to train their own algorithms and compare results with this implementation.

\begin{figure}[t]
\centering
\begin{subfigure}[t]{\columnwidth}
    \includegraphics[width=\linewidth]{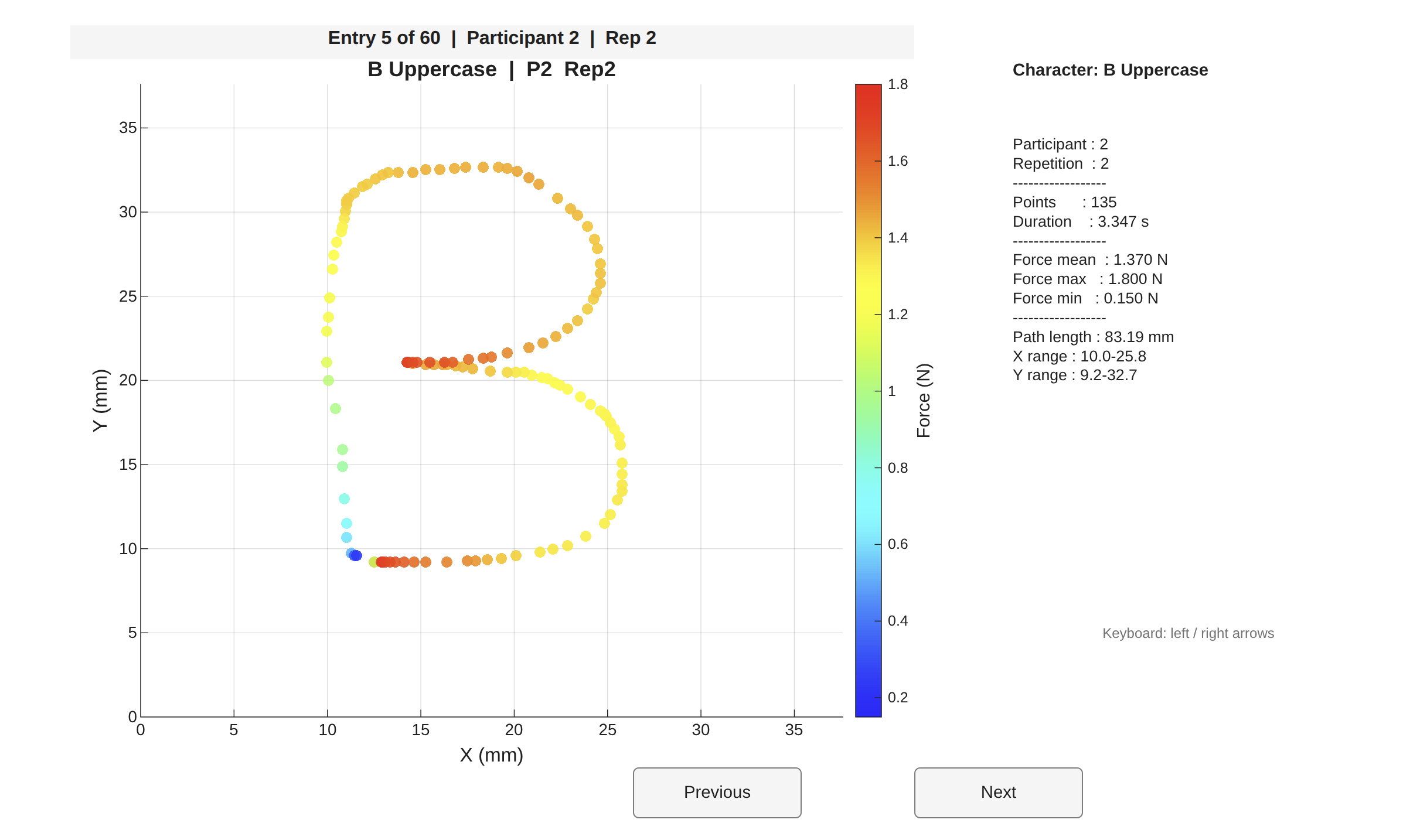}
    \caption{Top-down view of the recorded trajectory with force colour map.}
    \label{fig:2D}
\end{subfigure}

\begin{subfigure}[t]{\columnwidth}
    \includegraphics[width=\linewidth]{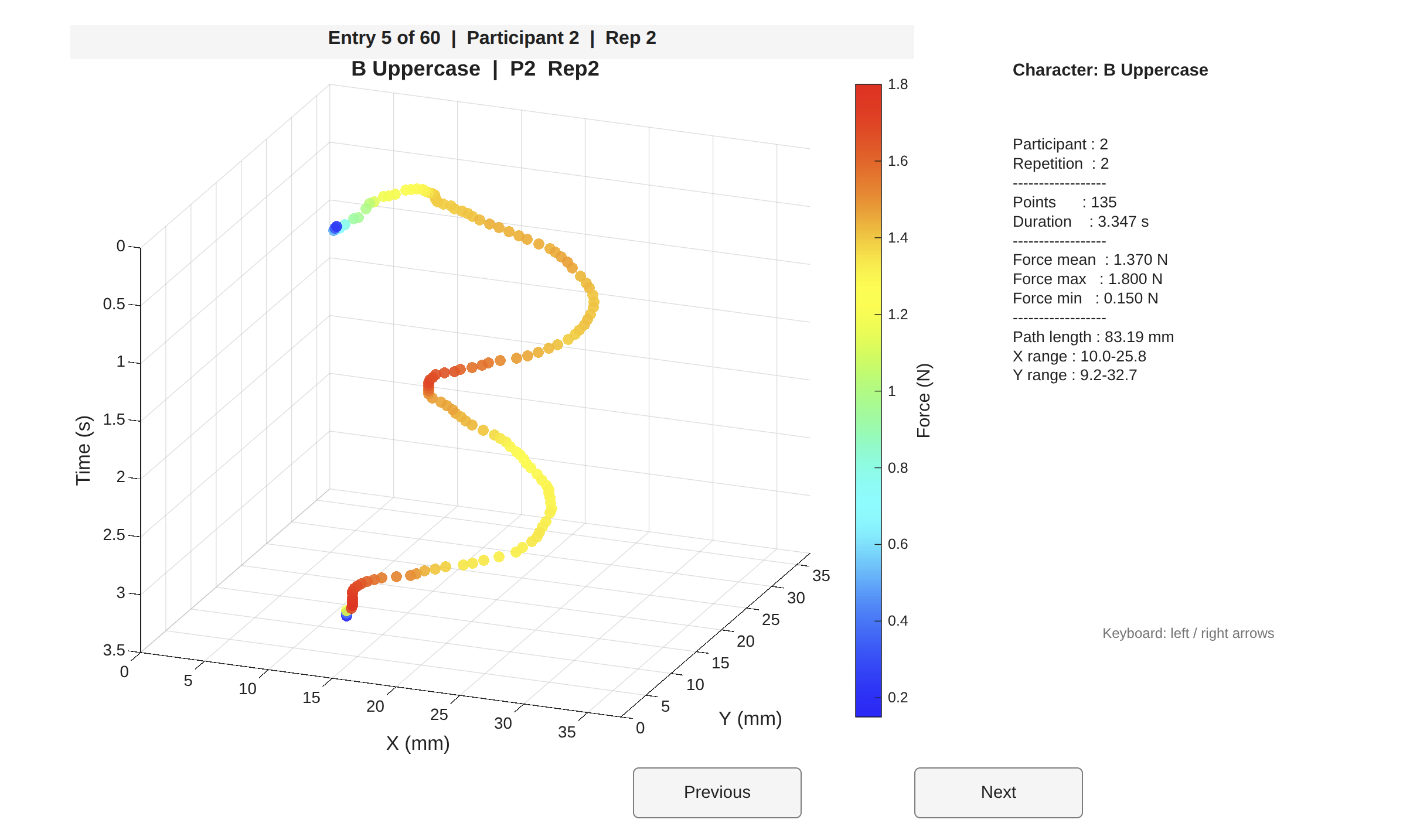}
    \caption{3D view of the recorded trajectory with time and force details.}
    \label{fig:3D}
\end{subfigure}
\caption{An example recorded human trajectory for uppercase B, second repetition by participant two.}
\label{fig:trajectory}
\end{figure}

Figure \ref{fig:trajectory} shows 2D and 3D views of an example trajectory for uppercase B, second repetition by participant two. The x and y axes show position in millimetres, the z axis represents the timestamp in seconds, and the colour map displays contact force in Newtons, with blue indicating lower and red higher force.

\subsection{Learning Algorithm}
\label{sec:Learning_Algorithm}

Robot motion learning from human demonstrations is implemented using GMM+GMR \cite{Calinon2012, Hewitt2017}. Two task-specific adaptations are introduced to meet the requirements of handwriting trajectory learning. First, the state vector is augmented beyond spatial coordinates to include contact force and normalised time as additional learned dimensions, enabling the robot to reproduce not only the geometry of handwriting but also its dynamics and contact behaviour. Second, a segment detection step is applied prior to training, which partitions each demonstration into individual strokes and trains each segment independently, addressing the trajectory degradation that arises at discontinuities caused by pen lifts between strokes.

\subsubsection{State Representation}

Each demonstrated trajectory is represented as a four-dimensional state vector 
$\mathbf{p} \in \mathbb{R}^4$, which stacks the quantities to be learned at each 
timestep:

\begin{equation}
    \mathbf{p}(t) = \begin{bmatrix} x(t) & y(t) & f(t) & t_{\text{norm}}(t) 
    \end{bmatrix}^\top
\end{equation}

where $x$ and $y$ are the pen tip positions in millimetres, $f$ is the contact force 
in Newtons, and $t_{\text{norm}} \in [0, 1]$ is the timestamp normalised by the 
average demonstration duration. This augmented representation allows the system to 
learn spatial, temporal, and force characteristics jointly from human demonstrations.

\subsubsection{GMM and GMR for Multi-Dimensional Trajectory Learning}

For each stroke segment, the training data is formed by combining the resampled 
state trajectories $\mathbf{p} \in \mathbb{R}^4$ across all $N$ demonstrations, 
paired with their corresponding timestamps $t \in \mathbb{R}$, where $t$ is a 
linearly spaced time vector spanning the average duration of the segment. Each 
training sample is therefore a five-dimensional vector 
$[t,\, x,\, y,\, f,\, t_{\text{norm}}]^\top$, and the full training set pools 
these samples across all $N$ demonstrations. GMM is used to encode the joint 
probability distribution over this combined time-state space:

\begin{equation}
    P(t, \mathbf{p}) = \sum_{k=1}^{K} \pi_k \, \mathcal{N}\!\left(
    \begin{bmatrix} t \\ \mathbf{p} \end{bmatrix}; \, \boldsymbol{\mu}_k, 
    \boldsymbol{\Sigma}_k\right)
    \label{eq:gmm}
\end{equation}

where $K$ is the number of Gaussian components, $\pi_k$ is the mixing coefficient 
of the $k$-th component satisfying $\sum_{k=1}^{K} \pi_k = 1$, and 
$\mathcal{N}(\cdot\,; \boldsymbol{\mu}_k, \boldsymbol{\Sigma}_k)$ denotes a 
multivariate Gaussian distribution with mean $\boldsymbol{\mu}_k$ and covariance 
$\boldsymbol{\Sigma}_k$. The mean vector is partitioned as 
$\boldsymbol{\mu}_k = [\mu_{t,k},\, \boldsymbol{\mu}_{\mathbf{p},k}]^\top$, where 
$\mu_{t,k} \in \mathbb{R}$ is the mean of the time dimension and 
$\boldsymbol{\mu}_{\mathbf{p},k} \in \mathbb{R}^4$ is the mean of the state 
dimensions for the $k$-th component. The covariance matrix is partitioned 
conformably as:

\begin{equation}
    \boldsymbol{\Sigma}_k = \begin{bmatrix} \Sigma_{tt,k} & 
    \boldsymbol{\Sigma}_{t\mathbf{p},k} \\ \boldsymbol{\Sigma}_{\mathbf{p}t,k} & 
    \boldsymbol{\Sigma}_{\mathbf{p}\mathbf{p},k} \end{bmatrix}
    \label{eq:sigma}
\end{equation}

where $\Sigma_{tt,k} \in \mathbb{R}$ is the variance of the time dimension, $\boldsymbol{\Sigma}_{\mathbf{p}\mathbf{p},k} \in \mathbb{R}^{4 \times 4}$ is the covariance among the state dimensions, and $\boldsymbol{\Sigma}_{\mathbf{p}t,k} = \boldsymbol{\Sigma}_{t\mathbf{p},k}^\top \in \mathbb{R}^{4 \times 1}$ is the cross-covariance between the state and time dimensions. All parameters are estimated via the Expectation-Maximisation (EM) algorithm \cite{Dempster1977} with $K = 20$ components, selected empirically based on trajectory reconstruction performance on held-out demonstrations. Once the GMM is fitted, time $t$ is used as the query dimension and the four state dimensions $[x,\, y,\, f,\, t_{\text{norm}}]$ are treated as response dimensions. GMR exploits the cross-covariance $\boldsymbol{\Sigma}_{\mathbf{p}t,k}$ 
to compute the conditional expectation of $\mathbf{p}$ given a query time $t$:

\begin{equation}
    \hat{\mathbf{p}}(t) = \sum_{k=1}^{K} h_k(t) \left[ \boldsymbol{\mu}_{\mathbf{p},k} 
    + \boldsymbol{\Sigma}_{\mathbf{p}t,k} \Sigma_{tt,k}^{-1} 
    \left(t - \mu_{t,k}\right) \right]
    \label{eq:gmr}
\end{equation}

where the term $\boldsymbol{\Sigma}_{\mathbf{p}t,k} \Sigma_{tt,k}^{-1}$ acts as a 
regression gain that adjusts the component mean $\boldsymbol{\mu}_{\mathbf{p},k}$ 
based on how far the query $t$ deviates from the component centre $\mu_{t,k}$, and 
$h_k(t)$ is the normalised activation weight of the $k$-th component:

\begin{equation}
    h_k(t) = \frac{\pi_k \, \mathcal{N}(t;\, \mu_{t,k},\, \Sigma_{tt,k})}
    {\sum_{j=1}^{K} \pi_j \, \mathcal{N}(t;\, \mu_{t,j},\, \Sigma_{tt,j}) + \epsilon}
    \label{eq:weights}
\end{equation}

where $\epsilon$ is a small regularisation constant added to the denominator to 
prevent division by zero when all components have negligible activation at a given 
query point. The regressed output $\hat{\mathbf{p}}(t)$ provides a smooth synthesised 
trajectory that represents the consensus across all $N$ demonstrations, with $x$, $y$, 
$f$, and $t_{\text{norm}}$ regressed jointly from a single time query. The reproduced trajectory therefore encodes not only the spatial path but also the expected contact force profile and the temporal pacing of the motion, both learned directly from human demonstrations. Technical details of the GMR formulation can be found in \cite{Calinon2012, Hewitt2017}.

\subsubsection{Segment-Wise Processing for Discontinuous Characters}

Many characters in the Latin alphabet require one or more pen lifts, introducing 
trajectory discontinuities that a single GMM cannot model without corrupting the 
learned distribution across strokes. To address this, a segment detection step 
identifies pen lifts in each demonstration by considering three cues: the Euclidean 
distance between consecutive position samples increases sharply at a pen lift, a 
temporal gap appears as time is spent lifting and repositioning the pen, and the 
contact force starts from near zero at the start of a new stroke. The primary detection criterion is defined as a threshold on the Euclidean distance in a four-dimensional feature space, where time and force are appropriately weighted to account for differences in units, computed between consecutive samples:

\begin{equation}
    \left\| \mathbf{x}_{i+1} - \mathbf{x}_i \right\|_2 > \delta_{\text{gap}}
    \label{eq:gap}
\end{equation}

where $\delta_{\text{gap}}$ is the threshold. Demonstrations are grouped by detected segment count, retaining those sharing the most common number to ensure training consistency. A separate GMM and GMR trajectory is fitted per segment independently, and segments are combined in temporal order with pen-lift intervals modelled as linear transitions. This unified approach handles both single and multi-stroke characters while preserving stroke ordering.

\begin{figure}[t]
    \centering
    
    \begin{subfigure}[t]{0.49\columnwidth}
        \centering
        \includegraphics[width=\linewidth]{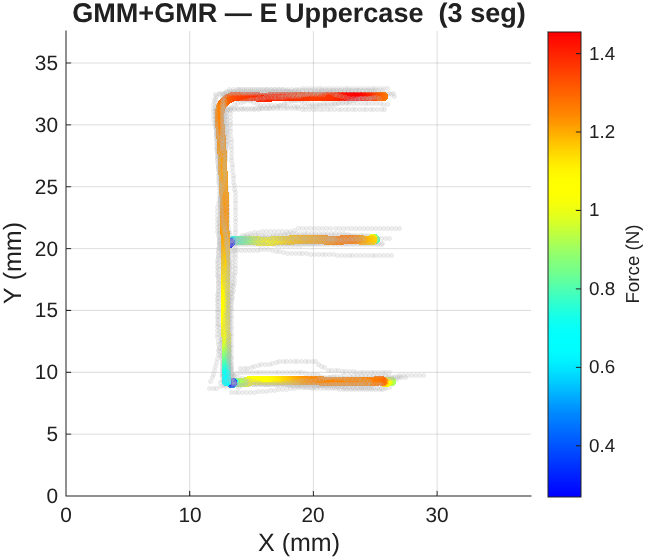}
        \caption{Top-down view of the generated trajectory}
        \label{fig:sub1}
    \end{subfigure}
    \hfill
    \begin{subfigure}[t]{0.49\columnwidth}
        \centering
        \includegraphics[width=\linewidth]{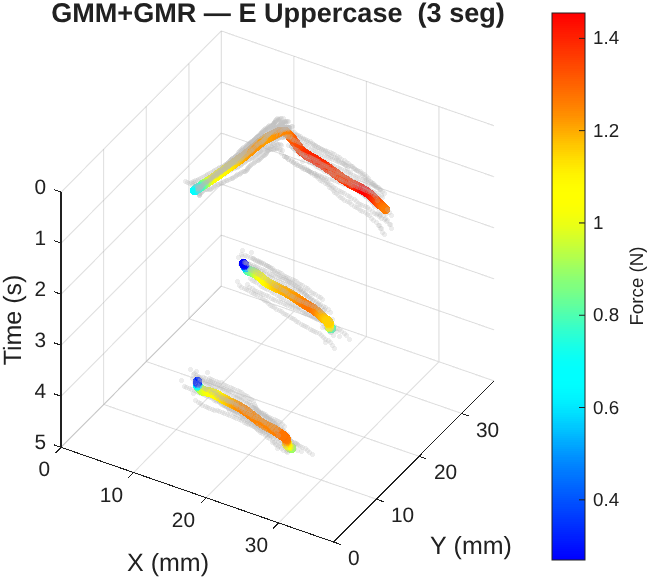}
        \caption{3D view of the generated trajectory}
        \label{fig:sub2}
    \end{subfigure}

    \caption{An example of a trajectory generated using the GMM+GMR approach, along with the training data visualised.}
    \label{fig:GMM}
\end{figure}

Figure \ref{fig:GMM} visualises a GMM+GMR application example, showing how the algorithm learns from multiple demonstration inputs to generate a single consensus trajectory, even for multi-segment paths with discontinuities. Grey points indicate the multiple demonstration data used for training.

\subsubsection{Output Trajectory}

Since $t_{\text{norm}}$ is learned as an additional state dimension rather than imposed on a uniform grid, the regressed time values in $\hat{\mathbf{p}}(t)$ may be unevenly spaced and can occasionally contain duplicate or out-of-order timestamps. These are corrected by rescaling to absolute time, sorting, and duplicate removal.

\begin{figure}[t]
    \centering
    
    \begin{subfigure}[t]{0.49\columnwidth}
        \centering
        \includegraphics[width=\linewidth]{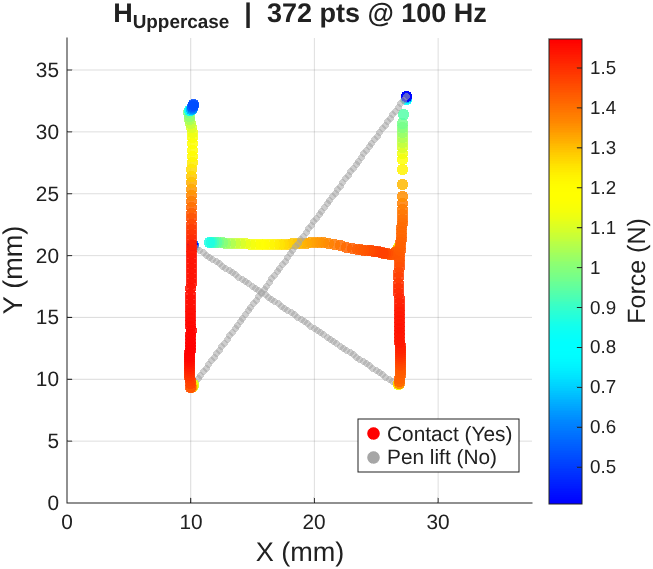}
        \caption{Top-down view of the output trajectory}
        \label{fig:sub1}
    \end{subfigure}
    \hfill
    \begin{subfigure}[t]{0.49\columnwidth}
        \centering
        \includegraphics[width=\linewidth]{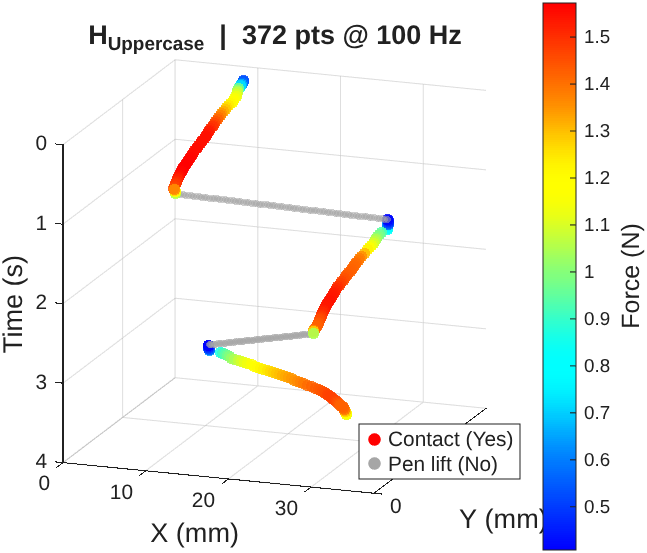}
        \caption{3D view of the output trajectory}
        \label{fig:sub2}
    \end{subfigure}
    
    \caption{Example of a generated trajectory compatible with ROS-based execution.}
    \label{fig:ROS}
\end{figure}

The corrected trajectory is subsequently resampled onto a uniform time grid at 100~Hz and exported as a CSV file. Each timestep contains the $x$ and $y$ position in millimetres, the contact force in Newtons, the absolute timestamp in seconds, and a binary contact flag indicating whether the pen is in contact with the surface. This output format is designed to be directly compatible with robot execution pipelines and is included in the repository\footnotemark[\value{repofootnote}] alongside the dataset. Figure \ref{fig:ROS} visualises an example output trajectory suitable for robotic applications. Grey-coloured points indicate segments where the pen is lifted, so there is no contact with the surface.

\subsection{Simulating Robot Motion}

In both studies, a UR10 manipulator was simulated in Unity 3D using an open-source robot model\footnote{Open-source UR10 robot model for Unity 3D simulation can be obtained from: \url{https://github.com/Preliy/Flange}}, with inverse kinematics 
used to command the end effector position. In the first study, the end effector followed 
the stylus position in real time through teleoperation. In the second study, it followed trajectories generated from the trained data. End effector speed was constrained and a small motion delay introduced to improve realism. In the second study, a marker attached to the end effector drew on a whiteboard, with trace thickness and transparency varying according to applied force.

\section{Results}
\label{sec:Results}

\subsection{Human Demonstration Dataset Details}

Demonstration trajectories capture the complete writing motion as a time series of planar position, normal contact force, and timestamp, recorded at 40~Hz. Table \ref{tab:demonstration_dataset} summarises the key statistical properties of the dataset computed across all valid trajectories. The spatial coverage across all demonstrations spans 3.53 - 36.11 mm along the $x$-axis and 0.00 - 37.03 mm along the $y$-axis, indicating effective utilisation of the available workspace. 

\begin{table}[h]
\centering
\caption{Statistical summary of the collected human demonstration dataset.}
\label{tab:demonstration_dataset}
\begin{tabular}{lcccc}
\hline
\textbf{Metric} & \textbf{Mean} & \textbf{Std} & \textbf{Min} & \textbf{Max} \\
\hline
Duration (s)        & 2.802  & 1.543 & 0.302  & 14.674  \\
Data points         & 95.5   & 49.5  & 13     & 360     \\
Force (N)           & 1.268  & 0.572 & 0.100  & 5.380   \\
Path length (mm)    & 65.31  & 20.38 & 19.91  & 128.45  \\
\hline
\end{tabular}
\end{table}

\subsection{Robot Learning Generated Trajectory Dataset Details}

The robot learning approach described in Section \ref{sec:Learning_Algorithm} was used to generate one trajectory per character-case combination, yielding a total of 52 trajectories. Each trajectory is stored as a time series of planar position, normal contact force, and timestamp, sampled at 100~Hz. The dataset is included in the project repository\footnotemark[\value{repofootnote}] alongside the human demonstration data for direct comparison or implementation. Table \ref{tab:rl_dataset} presents the key statistical properties of these generated trajectories. Mean contact activation represents the percentage of surface contact during each trajectory.

\begin{table}[h]
\centering
\caption{Statistical summary of the robot learning generated trajectory dataset.}
\label{tab:rl_dataset}
\begin{tabular}{lcccc}
\hline
\textbf{Metric} & \textbf{Mean} & \textbf{Std} & \textbf{Min} & \textbf{Max} \\
\hline
Duration (s)            & 2.702  & 0.744 & 1.030  & 4.390  \\
Data points             & 271.0  & 74.4  & 104    & 440    \\
Force (N)               & 1.084  & 0.257 & 0.000  & 1.868  \\
Path length (mm)        & 63.87  & 19.09 & 23.77  & 113.60 \\
Mean contact activation & 0.864  & 0.159 & --     & --     \\
\hline
\end{tabular}
\end{table}

\subsection{User Evaluation Results: Human-Likeness of Generated Trajectories}

To assess the perceived human-likeness of the robot-generated trajectories,
participants in the second user study evaluated each of the 52 character-case
combinations using a continuous Likert scale. The ratings were subsequently
linearly normalised to a 0-100 range, where higher values indicate greater
perceived human-likeness. In total, 1,092 ratings (26 characters $\times$ 2 cases $\times$ 21 participants) were collected. The overall mean human-likeness score was 71.50 ($SD = 22.56$). Overall, 887 out of 1092 ratings (81.2\%) exceeded the neutral midpoint of 50.

\begin{figure*}[t]
    \centering
    \includegraphics[width=\textwidth]{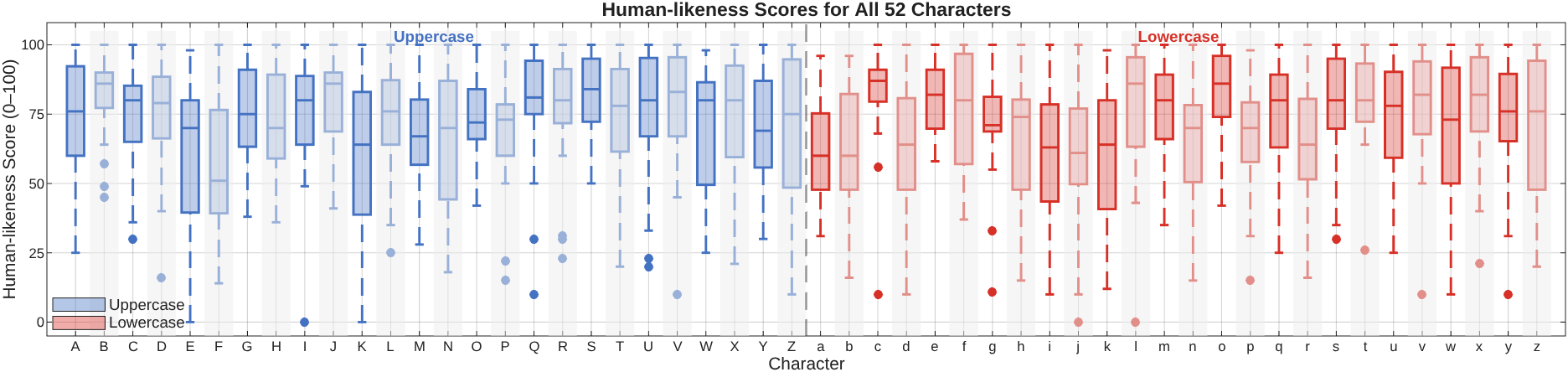}
    \caption{Human-likeness scores for all 52 character+case combinations. Each box
    represents the distribution of ratings (0-100) for that character. Uppercase
    are shown in blue, lowercase in red.}
    \label{fig:52char_boxplot}
\end{figure*}

Individual participant means ranged from 50.02 to 96.15. Uppercase characters received marginally higher ratings ($\bar{x} = 72.0$,
$SD = 22.4$) than their lowercase counterparts ($\bar{x} = 71.0$, $SD = 22.7$). Across all 52 character-case combinations, mean scores ranged from 57.43 to 83.95. The highest-rated combinations were \textit{o} (83.95), \textit{S} (82.38), and \textit{c} (81.76), while the lowest-rated were \textit{F} (57.43), \textit{j} (58.95), and \textit{k} (61.05). Standard deviations were generally high across both cases. The distribution of scores across all character-case combinations is shown in Figure \ref{fig:52char_boxplot}.

\subsection{Factors Influencing Human-Likeness}
Participants completed a post-study questionnaire rating the factors that influenced
their human-likeness judgements and the perceived impact of human-like robot motion
on collaboration, using a 5-point Likert scale (1 = strongly disagree,
5 = strongly agree). As shown in Table \ref{tab:q3_factors}, \textit{geometric positioning} and
\textit{trajectory sequence} were the most influential factors
($3.71 \pm 0.96$ and $3.57 \pm 1.03$, respectively), while \textit{applied
force} showed a similar level of influence ($3.52 \pm 1.03$) and
\textit{speed of motion} was the least influential ($2.81 \pm 1.08$).

\begin{table}[h]
\centering
\caption{Mean Likert ratings (1-5) for factors influencing human-likeness
judgements ($n=21$).}
\label{tab:q3_factors}
\begin{tabular}{lc}
\hline
\textbf{Factor} & \textbf{Mean $\pm$ SD} \\
\hline
Geometric positioning   & $3.71 \pm 0.96$ \\
Trajectory sequence     & $3.57 \pm 1.03$ \\
Applied force           & $3.52 \pm 1.03$ \\
Speed of motion         & $2.81 \pm 1.08$ \\
\hline
\end{tabular}
\end{table}

Participants reported generally positive attitudes toward human-like robot behaviour
(Table \ref{tab:q4_collab}), with \textit{preference for human-like behaviour} receiving the
highest agreement ($3.76 \pm 0.62$) and \textit{trust} the lowest ($3.24 \pm 0.94$).
\textit{Predictability} showed the greatest variability ($3.52 \pm 1.17$).

\begin{table}[h]
\centering
\caption{Mean Likert ratings (1-5) for collaboration-related statements ($n=21$).}
\label{tab:q4_collab}
\begin{tabular}{lc}
\hline
\textbf{Statement} & \textbf{Mean $\pm$ SD} \\
\hline
Preference for human-like behaviour & $3.76 \pm 0.62$ \\
Predictability                      & $3.52 \pm 1.17$ \\
Willingness to collaborate          & $3.43 \pm 0.87$ \\
Trust                               & $3.24 \pm 0.94$ \\
\hline
\end{tabular}
\end{table}

\section{Discussion}
\label{sec:Discussion}

The generated trajectories exhibit similar spatial coverage to the human demonstrations while showing more consistent durations and smoother force profiles, reflecting the regularising effect of the learning algorithm.

The human demonstration dataset reveals the natural variability inherent in human handwriting. Duration ranged from 0.302 to 14.674 s with a standard deviation of 1.543 s, and contact force varied from 0.100 to 5.380 N, reflecting the considerable individual differences across participants and repetitions. Despite this variability, the dataset collectively captures the spatial and dynamic characteristics of each character, providing a sufficiently rich basis for probabilistic learning. The similarity in mean path lengths between human demonstrations (65.31 mm) and robot-generated trajectories (63.87 mm) confirms that the GMM+GMR approach successfully preserves the spatial structure.

The overall mean human-likeness score of 71.50 ($SD = 22.56$), with 81.2\% of ratings exceeding the neutral midpoint, indicates that the learning approach succeeds in producing human-like trajectories. The high standard deviations reflect the subjective nature of human-likeness perception, consistent with \cite{dragan2013legibility}, who show that perceptual properties of robot motion are distinct and not always captured by a single measure. Geometric positioning and trajectory sequence were rated as the most influential factors, while speed of motion was rated least influential, indicating that timing mattered less than spatial characteristics to participants. Notably, applied force received a similarly high rating to trajectory sequence, which aligns with the force dimension explicitly modelled in the extended GMM+GMR framework.

Attitudes toward human-like robot behaviour were generally positive, with preference for human-like motion receiving the highest agreement ($3.76 \pm 0.62$), supporting the motivation for this work. These findings are broadly consistent with \cite{hostettler2023human}, who similarly found that human-like robot movement leads to more positive perceptions and stronger collaborative intent. The lower and more variable trust rating ($3.24 \pm 0.94$) suggests that human-likeness alone is not sufficient to fully establish trust, which is consistent with findings that trust in HRI is driven primarily by perceived robot performance rather than motion style alone \cite{hancock2011meta}.

This study makes three contributions to robot learning from demonstration. First, it provides a publicly available handwriting dataset capturing position, timing, and force data, offering a resource for benchmarking learning methods. Second, it extends the GMM+GMR framework with contact force and normalised time, enabling it to capture the full dynamics of human demonstrations beyond spatial paths and to generate complex multi-segment trajectories with discontinuities. Third, it presents a user evaluation of the generated trajectories using a simulated robot, providing empirical evidence of perceived human-likeness.

This work demonstrates a touchscreen-based interface for teaching a robot through physical demonstration without requiring programming expertise, lowering the barrier to accessible human-robot interaction. While validated on Latin alphabet characters, the pipeline is in principle applicable to any complex surface trajectory, suggesting broader deployment potential in tasks requiring human-like motion.

Limitations include the sample sizes of 22 and 21 participants and single laboratory settings for each study, which may limit generalisability. Free-space transitions between strokes were not captured, resulting in linear interpolation, and pen orientation was not recorded during teleoperation. The absence of a pre-programmed trajectory control condition limits direct comparison. Demonstrations followed predefined guidelines that may not reflect individual writing strategies, potentially limiting generalisation. More expressive writing styles, such as cursive or calligraphic writing, where pen orientation and force modulation play a greater role, remain for future work. Finally, the evaluation was conducted using a simulated robot rather than a physical system due to hardware access constraints, which may not fully capture real-world deployment conditions.

\section{Conclusion}
\label{sec:Conclusions}

This study presented a complete framework for teaching a robot human-like motion through demonstration. A dataset of 3,142 handwriting trajectories was collected from 22 participants across all 52 Latin alphabet character-case combinations, capturing position, timing, and force data via a touchscreen interface and a stylus pen. The GMM+GMR learning algorithm was extended with contact force and normalised time dimensions to generate smooth, multi-segment trajectories that preserve the spatial and dynamic characteristics of the human demonstrations. A user study with 21 participants evaluated the perceived human-likeness of the generated trajectories, yielding an overall mean score of 71.50 ($SD = 22.56$), with 81.2\% of ratings exceeding the neutral midpoint, indicating that the learning approach succeeds in producing trajectories perceived as human-like.

The results demonstrate that probabilistic LfD is effective for generating human-like trajectories, with the publicly available dataset and implementation providing an open-source benchmark resource.

\section*{Acknowledgement}

This work was supported by the European Commission’s Marie Skłodowska-Curie Actions (MSCA) Project RAICAM (GA 101072634), and UK Research and Innovation (UKRI) grant number EP/X025977/1. This work was inspired by coursework undertaken as part of the UFME7R-15-M Robot Learning \& Teleoperation module at UWE Bristol. 

For the purpose of open access, the author has applied a Creative Commons Attribution (CC BY) licence to any Author Accepted Manuscript version arising.

\bibliographystyle{IEEEtran}  
\bibliography{IEEEabrv}

\vspace{12pt}

\end{document}